\documentclass[format=sigconf,nonacm]{acmart}

\AtBeginDocument{%
  }

\usepackage{graphicx}
\usepackage{booktabs}
\usepackage{multirow}
\usepackage{colortbl}
\usepackage{adjustbox}
\usepackage{mathtools}
\usepackage{amsmath}
\usepackage{tabularx}
\usepackage{geometry}
\usepackage{enumitem}
\usepackage{pifont}
\usepackage{xcolor}
\usepackage{subcaption}
\usepackage[most]{tcolorbox}

\definecolor{E}{HTML}{689DF6}
\definecolor{L}{HTML}{815FA9}
\definecolor{S}{HTML}{A79F14}
\definecolor{F}{HTML}{56AC6D}
\definecolor{C}{HTML}{DB6156}

\definecolor{B}{HTML}{FB8315}
\definecolor{T}{HTML}{BAED6D}
\definecolor{SC}{HTML}{4FBDFE}

\begin{document}

\title{MPCC: A Novel Benchmark for Multimodal Planning with Complex Constraints in Multimodal Large Language Models}

\author{Yiyan Ji}
\authornote{Equal Contribution}
\email{jiyiiiyyy@gmail.com}
\affiliation{
  \institution{Harbin Institute of Technology}
  \city{Harbin}
  \state{Heilongjiang}
  \country{China}
}

\author{Haoran Chen}
\authornotemark[1]
\email{cshrchen@gmail.com}
\affiliation{
  \institution{Harbin Institute of Technology}
  \city{Harbin}
  \state{Heilongjiang}
  \country{China}
}

\author{Qiguang Chen}
\email{qgchen@ir.hit.edu.cn}
\affiliation{
  \institution{Harbin Institute of Technology}
  \city{Harbin}
  \state{Heilongjiang}
  \country{China}
  }

\author{Chengyue Wu}
\email{wcy010805@gmail.com}
\affiliation{
  \institution{The University of Hong Kong}
  \city{Hong Kong}
  \country{China}
}

\author{Libo Qin}
\authornote{Corresponding Author}
\email{lbqin@csu.edu.cn}
\affiliation{
 \institution{Central South University}
 \department{School of Computer Science and Engineering}
 \city{Changsha}
 \state{Hunan}
 \country{China}
 }

\author{Wanxiang Che}
\email{car@ir.hit.edu.cn}
\affiliation{
  \institution{Harbin Institute of Technology}
  \city{Harbin}
  \state{Heilongjiang}
  \country{China}
  }

\renewcommand{\shortauthors}{Yiyan Ji et al.}

\begin{abstract}
Multimodal planning capabilities refer to the ability to predict, reason, and design steps for task execution with multimodal context, which is essential for complex reasoning and decision-making across multiple steps. However, current benchmarks face two key challenges: (1) they cannot directly assess multimodal real-world planning capabilities, and (2) they lack constraints or implicit constraints across modalities. To address these issues, we introduce Multimodal Planning with Complex Constraints (MPCC), the first benchmark to systematically evaluate MLLMs' ability to handle multimodal constraints in planning.
To address the first challenge, MPCC focuses on three real-world tasks: Flight Planning, Calendar Planning, and Meeting Planning. To solve the second challenge, we introduce complex constraints (e.g. budget, temporal, and spatial) in these tasks, with graded difficulty levels (EASY, MEDIUM, HARD) to separate constraint complexity from search space expansion. Experiments on 13 advanced MLLMs reveal significant challenges: closed-source models achieve only 21.3\% feasible plans, while open-source models average below 11\%. Additionally, we observe that MLLMs are highly sensitive to constraint complexity and that traditional multimodal prompting strategies fail in multi-constraint scenarios.
Our work formalizes multimodal constraints in planning, provides a rigorous evaluation framework, and highlights the need for advancements in constraint-aware reasoning for real-world MLLM applications.
\end{abstract}


\maketitle

\section{Introduction}

Current multimodal large language models (MLLMs), such as Gemini~\cite{team2023gemini} and GPT-4o~\cite{Achiam2023GPT4TR}, demonstrate strong proficiency in processing diverse data modalities~\citep{ma2024janusflow,wu2024janus,wu2024deepseek}. They have been widely applied to tasks in multimodal understanding~\cite{yue2024mmmu} and reasoning~\cite{yang2023mmreactpromptingchatgptmultimodal,cheng2025visual,cheng2025comt}. On this basis, recent studies~\cite{ying2024mmt,xiao2024logicvista,ransford2011mementos,wang2024s3} have explored challenges in complex real-world planning.
To assess MLLMs' planning capabilities, benchmarks like PlanBench~\cite{valmeekam2023planbenchextensiblebenchmarkevaluating} provide diverse missions and targeted assessments. Similarly, m\&m's~\cite{ma2024m} offers tools specifically for MLLM planning. Other benchmarks, such as ALFRED~\cite{shridhar2020alfred} and Behavior-1K~\cite{li2023behavior}, evaluate embodied planning in simulated home environments using natural language instructions~\citep{lv2024robomp}. Additionally, WebArena~\cite{zhou2023webarena} and OSWorld~\cite{xie2024osworld} assess MLLM planning within websites and operating systems.

\begin{figure}[t]
  \begin{subfigure}{0.42\textwidth}
  \centering
  \includegraphics[width=\textwidth]{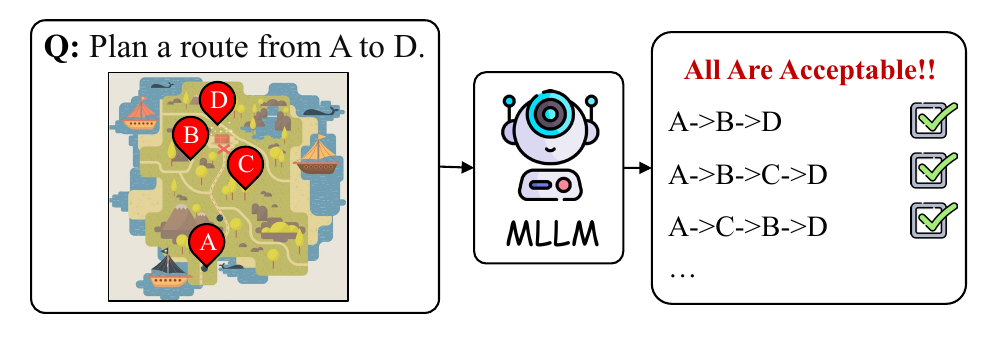}
  \caption{Multimodal planning task without any constraint.}
  \label{fig:introa}
\end{subfigure}
  \begin{subfigure}{0.42\textwidth}
  \centering
  \includegraphics[width=\textwidth]{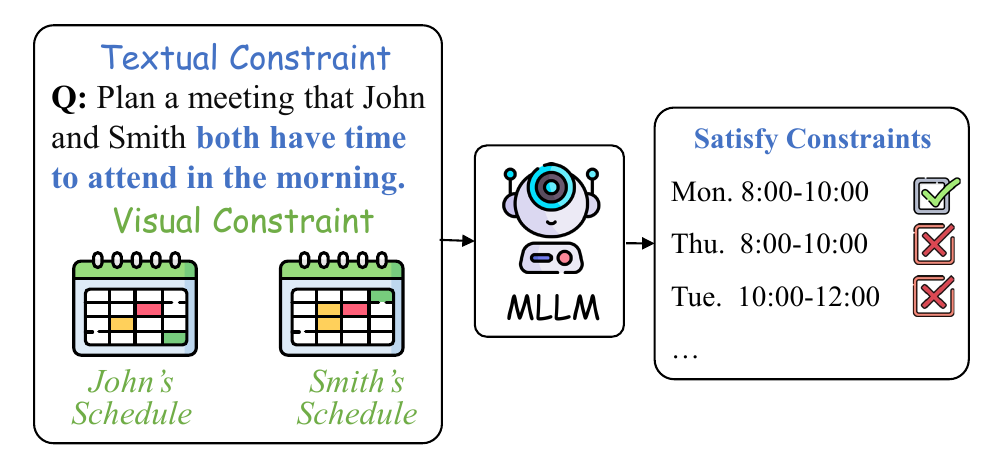}
  \caption{Multimodal planning task with multimodal constraints.}
  \label{fig:introb}
\end{subfigure}
  \caption{The example of multimodal planning tasks without any constraint (a) and with multimodal constraints (b).
  }
  \label{fig:intro}
\end{figure}

\begin{table*}[t]
  \centering
  \caption{Comparison of MPCC with other multimodal reasoning benchmarks. \textbf{\#Q} represents the size of questions; \textbf{\#I} represents the total number of images; \textbf{Cons.} represents different types of constraints.}
  
  \begin{adjustbox}{width=0.92\linewidth}
  \begin{tabular}{llcccccc}
  \toprule
  \multirow{1}{*}{\textbf{Dataset}} & \multirow{1}{*}{\textbf{Task Type}} & \multicolumn{1}{c}{\textbf{\#Q}} & \multicolumn{1}{c}{\textbf{\#I}} & \multicolumn{1}{c}{\textbf{Real-world}} & \multicolumn{1}{c}{\textbf{Graded Difficulties}} & \multicolumn{1}{c}{\textbf{Multi-modal Cons.}} & \multicolumn{1}{c}{\textbf{Composite Cons.}} \\ 
  
  \midrule    
  
  MMMU~\cite{yue2024mmmu} & Understanding & 11,550 & 11,264 & \textcolor{red}{\ding{55}} & \textcolor{red}{\ding{55}} & \textcolor{red}{\ding{55}} & \textcolor{red}{\ding{55}} \\
  SEED-Bench~\cite{li2023seed} & Generation & 19,242 & - & \textcolor{red}{\ding{55}} & \textcolor{red}{\ding{55}} & \textcolor{red}{\ding{55}} & \textcolor{red}{\ding{55}} \\
  MLLM-CompBench~\cite{kil2024mllm} & Reasoning & 39,800 & 79,600 & \textcolor{red}{\ding{55}} & \textcolor{red}{\ding{55}} &\textcolor{red}{\ding{55}} & \textcolor{red}{\ding{55}} \\
  EgoPlan-Bench~\cite{chen2023egoplan} & Planning & 4,939 & - & \textcolor{green}{\ding{51}} & \textcolor{red}{\ding{55}} & \textcolor{red}{\ding{55}} & \textcolor{red}{\ding{55}} \\
  \midrule
  \textbf{MPCC}(ours) & Planning & 2,700 & 6,300 & \textcolor{green}{\ding{51}} & \textcolor{green}{\ding{51}} & \textcolor{green}{\ding{51}} & \textcolor{green}{\ding{51}} \\
  
  \bottomrule
  \label{tab:comaprison}
  \end{tabular}
  \end{adjustbox}
\end{table*}

Despite significant advancements in multimodal planning, current benchmarks primarily focus on expanding the planning search space~\cite{koh2024treesearchlanguagemodel, shridhar2020alfred, li2023behavior, zheng2024naturalplanbenchmarkingllms, valmeekam2023planbenchextensiblebenchmarkevaluating} or improving complex multimodal perception~\cite{li2024ferretui2masteringuniversal, wang2025xlrsbenchmultimodalllmsunderstand, patraucean2023perception, ying2024mmt}. However, real-world planning complexity stems from satisfying intricate constraints across modalities. In the absence of constraints, almost any outcome could be deemed acceptable. For example, when planning a route from A to D (Figure~\ref{fig:intro} (a)), multiple paths can lead to the same goal, but only those that meet specific requirements are viable. A more critical challenge is managing these constraints across various factors. As shown in Figure~\ref{fig:intro} (b), real-world planning tasks, such as coordinating schedules for multiple attendees, require careful consideration of spatial and temporal constraints to generate a feasible plan.
Unfortunately, current benchmarks do not adequately address the resolution of practical and composite multimodal constraints, which are essential for reliable real-world planning with MLLMs.

To address this gap, we propose a novel benchmark, \textbf{Multimodal Planning with Complex Constraints (MPCC)}, designed to systematically evaluate the planning capabilities of MLLMs under complex multimodal constraints. Specifically, our dataset is constructed from real-world scenarios, generated using a code generator, and then manually filtered to ensure accurate constraint specifications and sufficient combinatorial diversity.
For practical applications, \textbf{MPCC} encompasses three widely-used planning tasks, each involving progressively complex constraints: Flight Planning, Calendar Planning, and Meeting Planning.
To more effectively assess planning capabilities, each task is categorized into three difficulty levels, while ensuring a comprehensive evaluation of both the complexity of constraints and the challenges with search space expansion.

Our evaluation reveals that \textbf{MPCC} presents a significant challenge for state-of-the-art MLLMs. In the Easy Meeting Planning task, the most advanced model, Claude-3.5V-Sonnet, achieves only 46.7\% optimal plans. As task complexity and constraints increase, model performance declines by more than 10\%. For tasks with higher constraint levels, model performance often drops below random chance. We also evaluated conventional optimization methods, including In-Context Learning (ICL)~\cite{dong2022survey} and Chain of Thought (CoT)~\cite{kojima2023largelanguagemodelszeroshot}, which yielded several key insights: (1) \textit{Increasing constraint complexity reduces the effectiveness of CoT reasoning}; (2) \textit{ICL significantly interferes with multimodal planning processes.}

In conclusion, our key contributions are summarized as follows:
\begin{itemize}[topsep=1ex,leftmargin=4ex,itemsep=1ex]
  \item We propose the concept of "multimodal constraint" for the first time in the context of multimodal planning problems and propose a novel benchmark, Multimodal Planning with Complex Constraints (MPCC), for more effective evaluation of planning tasks in multimodal environments.
  \item We demonstrate that MPCC represents a highly challenging benchmark, highlighting the difficulties that existing multimodal models face when handling planning tasks with increased constraints. This observation encourages researchers to reconsider the development of MLLMs.
  \item We conduct tests using conventional optimization methods and analyze various factors that influence the performance of MLLMs in complex constrained planning tasks, with the aim of inspiring further advancements in this area.
\end{itemize}

To facilitate further research, resources are available at \url{https://github.com/j-yyyyy/MPCC}.

\section{Multimodal Constraints}
\label{sec:constraints}

\begin{figure*}[t]
  \centering
    \includegraphics[width=0.98\textwidth]{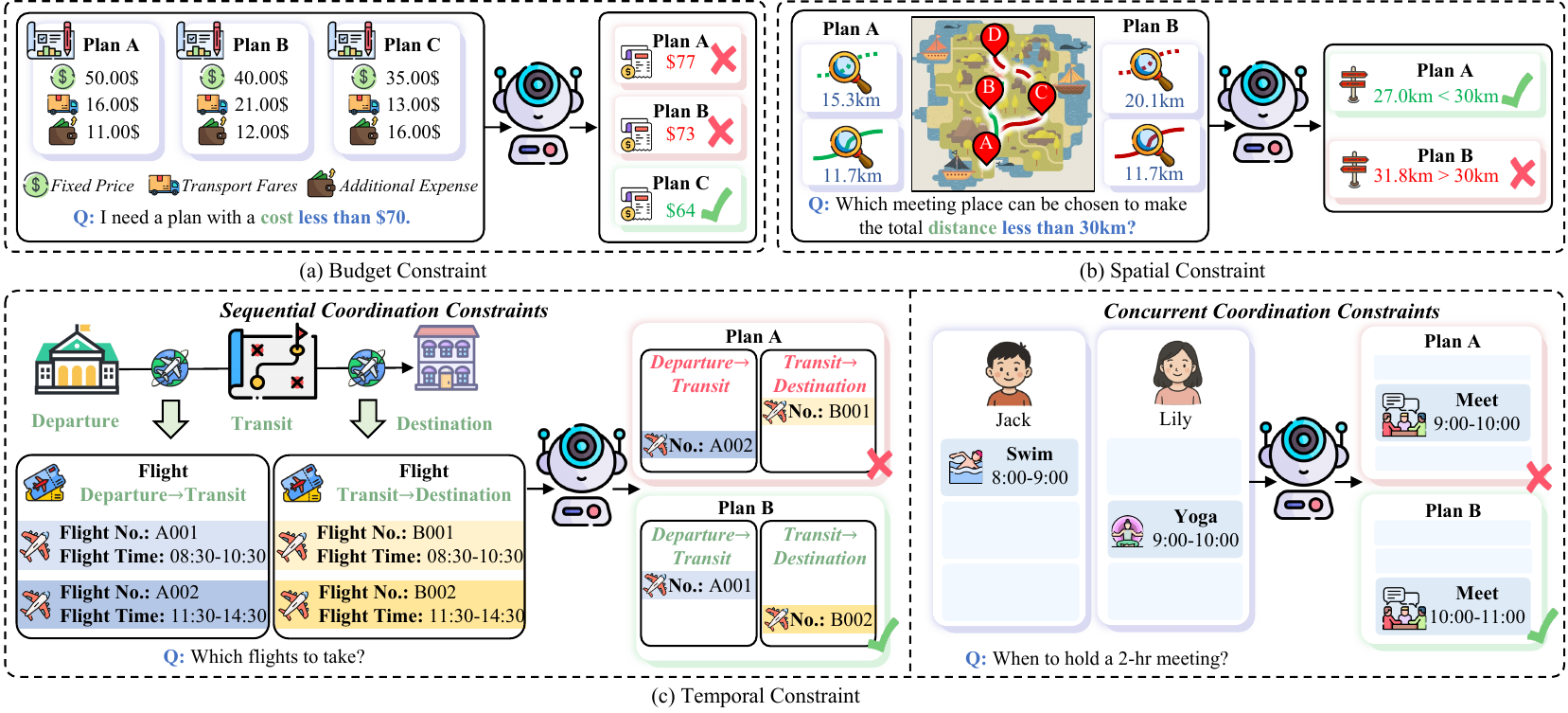}
    \caption{An overview of the categories of multimodal constraints featured in MPCC. These constraints are instantiated through both visual and textual inputs, demanding MLLMs to perform joint reasoning across modalities. MPCC adopts these realistic constraint categories to systematically evaluate models' abilities in planning under complex, multimodal scenarios.}
    \label{fig:constraints}
  \end{figure*}

In this benchmark, we introduce three fundamental constraint categories that assess the planning capabilities of MLLMs.

\subsection{Budget Constraints ($\mathcal{B}$)}
As shown in Figure~\ref{fig:constraints} (a), budget constraints govern the total allocation of resources within a given plan. These constraints ensure that the solution adheres to a predefined budget limit, a common requirement in numerous real-world tasks. Specifically, the budget constraint dictates that the plan, $p$ must satisfy the condition:
\begin{equation}
\mathcal{B}: \sum_{j=1}^m cost(r_j|p) \leq B_{\text{max}},
\end{equation}
where $r_j$ denotes the booked resources (e.g., flight tickets or meeting rooms), $B_{max}$ represents the maximum allowable budget for the task, and $cost(r_j|p)$ refers to the expenditure associated with the resource $r_j$ under the plan $p$. A plan is deemed valid if the total expenditure does not exceed the predefined budget.

\subsection{Temporal Constraints ($\mathcal{T}$)}
Temporal constraints ensure that all activities in the plan are completed within a feasible time frame. These constraints are enforced through two primary mechanisms:

\paragraph{Sequential Coordination}: 
As illustrated in Figure~\ref{fig:constraints} (c), sequential coordination applies when the order of events is critical, such as in flight connections, where the time between consecutive flights must fall within an allowable range. Formally, this is expressed as:
\begin{equation}
\mathcal{T}_{\text{seq}}: \sum_{i=1}^k \Delta t_i \leq T_{\text{max}}, \Delta t_i \ge t_{\text{min}},
\end{equation}
where $\Delta t_i$ represents the buffer time between consecutive flights, $t_{\text{min}}$ is the minimum interval between events, and $T_{\text{max}}$ is the maximum allowable cumulative time for the tasks. This ensures that sufficient time is allocated for each leg of the journey, including waiting periods at airports.

\paragraph{Concurrent Coordination}: 
Concurrent coordination ensures that all participants are available at the same time for a particular activity (see Figure~\ref{fig:constraints} (c)), such as scheduling a meeting. Formally, this is represented as:
\begin{equation}
    \mathcal{T}_{\text{con}} : \bigcap_{i=1}^n A_i \neq \varnothing,
\end{equation}
where $A_i$ denotes participant $i$-th available time slots, and the intersection of all these sets must be non-empty. This guarantees that there is at least one time slot during which all participants can attend the meeting or event.

\subsection{Spatial Constraints ($\mathcal{S}$)}
Spatial constraints address the feasibility of spatial distances as shown in Figure~\ref{fig:constraints} (b). These constraints are relevant when the locations of participants or resources must be considered. For instance, in meeting planning, the distance between meeting locations and participants' locations must not exceed a specified maximum travel time. Formally, this is represented as:
\begin{equation}
\mathcal{S} : d(l(p_i), l_{\text{meet}}) \leq D_{i},
\end{equation}
where $l_{\text{meet}}$ denotes the spatial location, and $p_i$ represents participant $i$-th location. $D_{i}$ is the maximum distance participant $i$ can travel, ensuring that no participant needs to travel an unreasonable distance to attend an activity.

\section{Multimodal Planning Complex Constraint (MPCC) Benchmark}

We introduce the Multimodal Planning Complex Constraint (MPCC) benchmark, which is designed to evaluate the ability of multimodal models to extract information and perform planning tasks under varying types of constraints. The comparison of our dataset with other multimodal datasets is shown in Table~\ref{tab:comaprison}. Referring to NATURAL PLAN~\cite{zheng2024naturalplanbenchmarkingllms}, our dataset contains three task categories: Flight Planning, Calendar Planning, and Meeting Planning. Each type of planning task consists of "EASY", "MEDIUM", and "HARD" levels of difficulty to ensure a variety of problem formats and search spaces.

\subsection{Composite Constraint Construction}
To construct complex interaction scenarios involving multiple composite constraints, we design three representative multimodal planning tasks grounded in real-world scenarios:

\subsubsection{Flight Planning}
\label{subsec:flight}
In flight planning, a common scenario is the absence of direct flights between two cities. Here, MLLMs help users select connecting flights and generate a complete itinerary. The composite constraint $\mathcal{C}$ combines a sequential temporal constraint $\mathcal{T}_{seq}$, limiting transfer intervals, and a budget constraint $\mathcal{B}$, restricting total cost. Formally, it is defined as:
\begin{equation}
  \mathcal{C} = \mathcal{T}_{\text{seq}} \otimes \mathcal{B}
\end{equation}
where $\otimes$ denotes the joint enforcement of constraints. The interface follows Google Flights, with all flight data synthetically generated and anonymized. Users' flight schedules ($\mathcal{T}_{seq}$) and budgets ($\mathcal{B}$) are randomly assigned, ensuring at least one feasible solution exists.

\subsubsection{Calendar Planning}
\label{subsec:calendar}
Calendar Planning tackles scheduling multi-participant meetings under dynamic constraints. MLLMs must reconcile participants' calendars with time, room availability, and costs. The composite constraint $\mathcal{C}$ combines a temporal coordination constraint $\mathcal{T}_{con}$ and a budget constraint $\mathcal{B}$, defined as:
\begin{equation}
  \mathcal{C} = \mathcal{T}_{\text{con}} \otimes \mathcal{B}.
\end{equation}
The interface simulates Google Calendar with a synthesized schedule of events and time slots, reflecting the temporal coordination constraint $\mathcal{T}_{con}$. To incorporate the budget constraint $\mathcal{B}$, meeting rooms are assigned booking prices, and a randomly generated budget is introduced, which the meeting plan must satisfy.

\subsubsection{Meeting Planning}
\label{subsec:meeting}
Meeting Planning simulates real-world meeting organization and further expands composite constraints by introducing spatial constraints. The model must arrange meeting time and location while satisfying participant schedules, spatial distribution, and budget limits.
The composite constraint $\mathcal{C}$ combines spatial constraints $\mathcal{S}$, temporal coordination constraints $\mathcal{T}_{\text{con}}$, and budget constraints $\mathcal{B}$, formally defined as:

\begin{equation}
  \mathcal{C} = \mathcal{T}_{\text{con}} \otimes \mathcal{S} \otimes \mathcal{B}.
\end{equation}

This task uses a Google Maps-style interface to display randomly generated spatial backgrounds, participant locations, and potential meeting venues, each with corresponding distances. It also provides local travel speeds and transportation costs to reflect the spatial constraint $\mathcal{S}$. In the Meeting Planning task, participants' schedules are set up similarly to Calendar Planning, but planners must also factor in round-trip travel time when selecting a meeting time. Additionally, each venue has a unique booking cost, requiring planners to consider both travel expenses and the venue fee. This setup combines spatial constraint $\mathcal{S}$, temporal coordination constraint $\mathcal{T}_{con}$, and budget constraint $\mathcal{B}$, resulting in a realistic planning scenario.

\subsection{Constraint Complexity Construction}

To rigorously assess multimodal planning under increasing constraint complexity, we adopt a progressive evaluation scheme based on the concept of search space, which includes all valid plans under a task's conditions. As task complexity increases, so do the variables and their combinations, making reasoning and optimization more difficult. Rather than arbitrarily scaling constraints, we adjust scenario parameters that influence the plan space.
In Flight Planning, complexity depends on the number of transit cities, planning days, and flight options per route.
In Calendar and Meeting Planning, it depends on planning duration, venue availability, and schedule resolution. These adjustments are calibrated to ensure that valid configurations grow consistently across tasks and difficulty levels with comparable search space sizes and task-specific semantics.

To generate an optimal plan, a brute-force search is performed for each task instance to explore all possible solutions within the constraints of budget, time, and space. Infeasible plans are discarded, and the optimal solution is selected. Instances without feasible solutions are eliminated, ensuring every task has a definitive solution.

\subsection{Multimodal Constraints Ensurement}
To validate and enhance the challenge of multimodal constraints in the MPCC dataset, we employ a systematic data construction process, as shown in Figure~\ref{fig:constraints}. Initially, interface frames from real-world applications (e.g., Google Flights \& Calendar) are collected and used to generate diverse images through randomization states, serving as multimodal inputs.
To ensure each example meets both visual and textual constraints, we apply human pre-labeling to filter out cases where the optimal solution is clear from a single modality, enforcing reliance on both modalities for optimal planning.

\subsection{Human Recheck}
To ensure the reliability of our dataset, we conduct a two-stage human recheck involving three experts with backgrounds in NLP and multimodal reasoning. The first stage checks whether visual and textual information aligns with the defined constraints; the second verifies that each instance admits at least one valid solution. Annotators follow standardized guidelines covering multimodal alignment, constraint consistency, and semantic clarity.
Each instance is reviewed independently by the annotators. Disagreements are resolved by majority vote or group discussion. The inter-annotator agreement (Kappa = 0.83) indicates strong consistency. Less than 10\% of auto-generated instances are discarded due to issues like ambiguous constraints or modality mismatch. This process ensures the dataset's high quality and alignment with task objectives.

\section{Metric Design}
In real-world planning tasks, multiple feasible plans often exist. To avoid overly rigid criteria for solution evaluation, we formalize two categories of constrained plans.

\paragraph{Feasible Plan Rate}
A feasible plan specifies the conditions that any valid plan must meet. In MPCC, these include: (1) upper bounds on total expenditure (budget ceilings), (2) temporal coordination requirements, and (3) spatial proximity limits (distance thresholds). A plan satisfying all such requirements is considered a \textit{feasible plan}. The rate at which MLLMs generate feasible plans indicates their ability to operate under complex constraints.

\paragraph{Optimal Plan Rate}
While humans can easily select the best option from feasible alternatives, this is more challenging for MLLMs. To assess their ability to explore and filter solutions, we use the budget as the optimization objective. A feasible plan minimizing budget is considered an \textit{optimal plan}. The rate at which MLLMs generate such plans indicates their capacity for optimal planning under complex constraints.

\begin{table}[t]
  \centering
  \caption{Detailed statistics of our dataset.}
  \begin{adjustbox}{width=0.30\textwidth}
	\begin{tabular}{l c}
		\toprule
		\textbf{Statistic} & \textbf{Number} \\
		\midrule
		Flight Planning size & 900 \\
		Calendar Planning size & 900 \\
		Meeting Planning size & 900 \\
		\midrule
		EASY level size each domain & 300 \\
		MEDIUM level size each domain & 300 \\
		HARD level size each domain & 300 \\
		\midrule
		EASY average search space & 27 \\
		MEDIUM average search space & 184 \\
		HARD average search space & 617 \\
		\midrule
		Average feasible plans account & 38.09 \\
		\bottomrule
		\end{tabular}
  \end{adjustbox}
  
  \label{tab:dataset}
  \end{table}
\subsection{Data Analysis}

\subsubsection{Basic statistics}
Our benchmark consists of 2.7K planning tasks, divided into three categories: Flight Planning, Calendar Planning, and Meeting Planning, with 0.9K instances per category. Each task type is categorized into three difficulty levels.
\subsubsection{Search Space Analysis}
To differentiate the dataset, we varied the search space size across levels. As shown in Table~\ref{tab:dataset}, the average search space size increases from 27 in the Easy level to 617 in the Hard level, representing a 22-fold increase. This structured expansion allows us to assess model performance under increasing complexity.
\subsubsection{Sparse Feasible Regions}
\begin{figure}[b]
      \centering
      \includegraphics[width=0.45\textwidth]{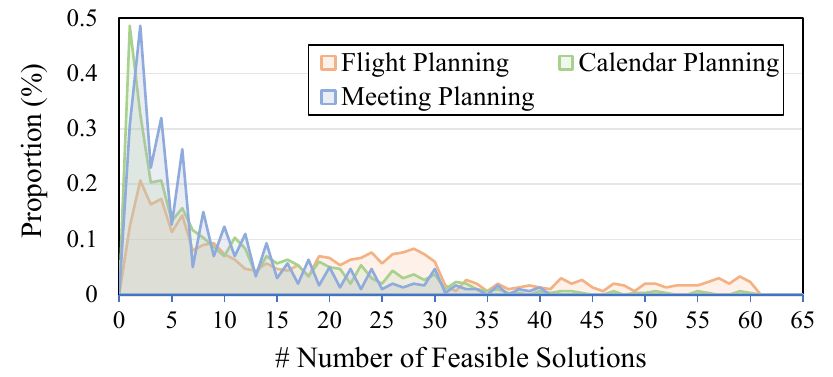}
      \caption{Distribution of the number of \textit{feasible plans} across different tasks.}
      \label{fig:feasible}
  \end{figure}

Figure~\ref{fig:feasible} shows the distribution of \textit{feasible plans} within the search space. While the average number of feasible solutions is 38.09, the distribution is highly skewed across tasks and difficulty levels, indicating that feasible plans represent a small fraction of the total search space. This highlights the importance of efficient reasoning and constraint satisfaction, as models must navigate large search spaces with few feasible solutions.

\begin{table*}[t]
	\centering
	\caption{
        Accuracy on \textit{feasible plan/optimal plan} in evaluations of selected MLLMs. \textbf{Empirical Max} stands for the maximum accuracy attainable by exhaustively evaluating all options within the search space, where each option is uniformly applied to all tasks for a given difficulty level, MLLMs that scored below this value were not considered capable of solving the tasks.
	}
	\begin{adjustbox}{width=0.96\textwidth}
		\begin{tabular}{lcccccccccc}
			\toprule
			\multirow{2}{*}{Model}  & \multicolumn{3}{c}{Flight Planning} & \multicolumn{3}{c}{Calendar Planning} & \multicolumn{3}{c}{Meeting Planning} & \multirow{2}{*}{Average}
			\\\cmidrule{2-10}
			& EASY & MEDIUM & HARD & EASY & MEDIUM & HARD & EASY & MEDIUM & HARD & 
			\\
			\midrule
			Empirical Max & - & - & - & 11.7/10.0 & \textbf{11.3}/\textbf{10.3} & \textbf{7.0}/\textbf{5.3} & 12.0/8.0 & \textbf{8.0}/5.3 & \textbf{7.0}/\textbf{6.3} & - \\
			\midrule
			\rowcolor{gray!8}\multicolumn{11}{c}{\textit{Closed-source MLLMs}}\\
			\midrule
			{GPT-4o} & 65.0/33.0 & 16.7/1.7 & 2.7/0.0 & 24.0/21.0 & 9.0/7.7 & 2.0/1.0 & 17.7/16.0 & 7.7/5.3 & 4.0/3.3 & 16.7/9.9 \\
			{Gemini-2.0-Flash} & 53.3/27.0 & 11.7/1.0 & 0.7/0.0 & 21.0/12.0 & 5.7/1.7 & 3.3/1.0 & 25.0/23.0 & 6.7/4.3 & 4.0/2.7 & 14.6/8.1 \\
			{Claude-3.5V-Sonnet} & \textbf{75.3}/\textbf{46.7} & \textbf{34.0}/\textbf{4.7} & 0.0/0.0 & \textbf{29.0}/\textbf{23.3} & 9.3/2.0 & 2.0/1.7 & \textbf{37.7}/\textbf{34.3} & 2.7/1.7 & 2.0/1.7 & \textbf{21.3}/\textbf{12.9} \\
			\midrule
			\rowcolor{gray!8}\multicolumn{11}{c}{\textit{Open-source MLLMs}}\\
			\midrule
            {InternVL-4B} & 18.0/3.3 & 0.0/0.0 & 0.0/0.0 & 4.3/3.3 & 7.0/0.3 & 2.0/2.0 & 9.0/5.3 & 6.3/5.0 & 4.3/1.0 & 5.7/2.2 \\
    		{InternVL-8B} & 32.0/11.0 & 0.0/0.0 & 0.0/0.0 & 10.7/7.7 & 0.3/0.0 & 4.0/0.0 & 5.3/4.0 & 1.3/1.3 & 1.0/0.7 & 6.1/2.7 \\
            {InternVL-26B} & 31.7/10.0 & 1.3/0.3 & 0.3/0.0 & 13.7/7.3 & 5.0/0.7 & 5.7/2.0 & 11.3/10.0 & \textbf{8.0}/\textbf{7.0} & 4.3/4.0 & 9.0/4.6 \\
            {InternVL-38B} & 35.7/13.0 & 0.7/0.3 & 0.0/0.0 & 14.0/9.0 & 5.7/0.0 & 3.3/3.0 & 6.3/5.3 & \textbf{8.0}/6.7 & 3.0/1.7 & 8.5/4.3 \\
            {InternVL-78B} & 37.3/15.0 & 1.7/0.0 & 0.3/\textbf{0.3} & 7.3/3.3 & 5.7/1.0 & 3.0/3.0 & 7.3/6.3 & 7.7/6.7 & 3.0/2.7 & 8.1/4.3 \\
            {LLaVa-OV-0.5B} & 0.0/0.0 & 0.0/0.0 & 0.0/0.0 & 3.7/3.0 & 1.7/0.0 & 1.3/1.3 & 6.3/3.0 & 0.3/0.3 & 1.0/0.7 & 1.6/0.9 \\
            {LLaVa-OV-7B} & 6.7/2.7 & 0.0/0.0 & 0.0/0.0 & 9.0/6.3 & 5.7/0.0 & 2.3/0.0 & 7.7/4.0 & 7.3/4.0 & 2.3/1.3 & 4.6/2.0 \\
            {Qwen2-VL-2B} & 5.3/1.0 & 0.0/0.0 & 0.0/0.0 & 4.7/4.3 & 1.7/0.0 & 0.7/0.0 & 5.7/2.7 & 3.0/1.3 & 1.3/0.7 & 2.5/1.1 \\
            {Qwen2-VL-7B} & 32.0/8.7 & 0.0/0.0 & 0.0/0.0 & 12.7/9.0 & 7.3/0.0 & 6.3/0.3 & 9.0/4.0 & 7.0/3.7 & 4.7/3.0 & 8.8/3.2 \\
            {Qwen2-VL-72B} & 46.0/17.7 & 12.7/1.0 & \textbf{9.3}/\textbf{0.3} & 8.3/6.7 & 4.7/0.0 & 2.3/2.3 & 10.0/8.7 & 3.3/2.7 & 1.7/1.7 & 10.9/4.6 \\
			\bottomrule
		\end{tabular}
	\end{adjustbox}
	
	\label{main}
\end{table*}

\section{Experiments}

\begin{figure}[t]
    \centering
    \includegraphics[width=0.42\textwidth]{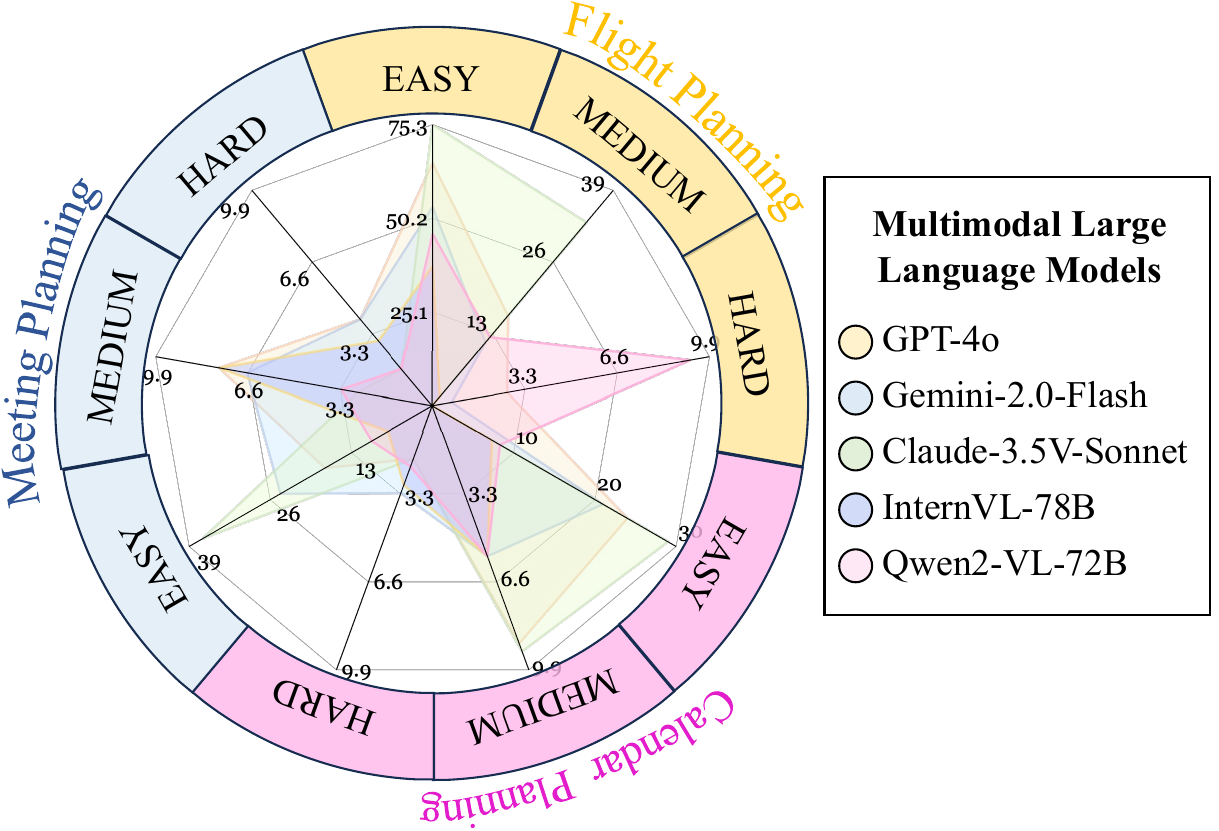}
    \caption{Performance of different models in \textit{feasible plan}.}
    \label{fig:radar}
\end{figure}

\subsection{Experiments Setting}
We evaluated a variety of state-of-the-art open- or closed-source MLLMs including GPT-4o~\cite{Achiam2023GPT4TR}, Gemini-2.0-Flash-EXP~\cite{team2023gemini}, Claude-3.5V-Sonnet, Qwen2-VL series~\cite{Qwen2-VL}, InternVL series~\cite{chen2024internvl} and LLaVa-OV series~\cite{li2024llavaonevisioneasyvisualtask}. We also explore different prompt strategies, including Chain-of-Thought~\cite{wei2022chain} and Plan-and-Solve~\cite{wang2023plan}. Our evaluations were conducted based on VLMEvalKit~\cite{duan2024vlmevalkit}, and the results were obtained through exact matching.

\subsection{Overall Evaluation}
The results are given in Table~\ref{main}, where we observe the following:

\paragraph{The MPCC can reveal significant challenges across tasks of varying difficulty}
As shown in Table~\ref{main}, performance drops significantly as task complexity increases. Closed-source models like GPT-4o show a sharp decline in Calendar Planning, from 24.0\% (EASY) to 2.0\% (HARD) in generating feasible plans, revealing strong sensitivity to growing constraints and search space. Even top closed-source models, such as Claude-3.5V-Sonnet (2.0\%), fail to outperform random selection. These results underscore the urgent need to improve reasoning under nonlinear search spaces and conflicting constraints, especially for open-source models, which still lag far behind.

\paragraph{There is still a substantial gap between open- and closed-source MLLMs for planning with complex multimodal constraints.}
Many MLLM reports suggest that open-source models perform comparably to closed-source ones in multimodal reasoning. However, as shown in Table~\ref{main}, even the most advanced open-source models fail to exceed 11.0\% average accuracy on MPCC tasks involving complex multimodal constraints—far below closed-source performance. The gap widens further as constraints increase. This highlights the need for open-source MLLMs to address constraint planning in multimodal scenarios better.

\subsection{Analysis}
\paragraph{The complexity of constraints significantly affects the planning ability of MLLMs}
As shown in Table~\ref{main}, mainstream MLLMs perform well on EASY-level Flight Planning, where constraints are simple. However, as illustrated in Figure~\ref{fig:radar}, their performance declines with increasing constraint complexity. For instance, most open-source models, regardless of scale, struggle in Calendar Planning tasks. The challenge intensifies in Meeting Planning, where even the best open-source model scores 0.7\% below the Empirical Max for \textit{feasible plan}, revealing their limitations in handling complex constraints. In contrast, among closed-source models, only Claude-3.5V-Sonnet actively uses Chain-of-Thought in Meeting Planning and achieves a relatively high score (37.7\%). Others show a clear decline as constraints increase.

\begin{figure}[t]
    \centering
    \includegraphics[width=0.45\textwidth]{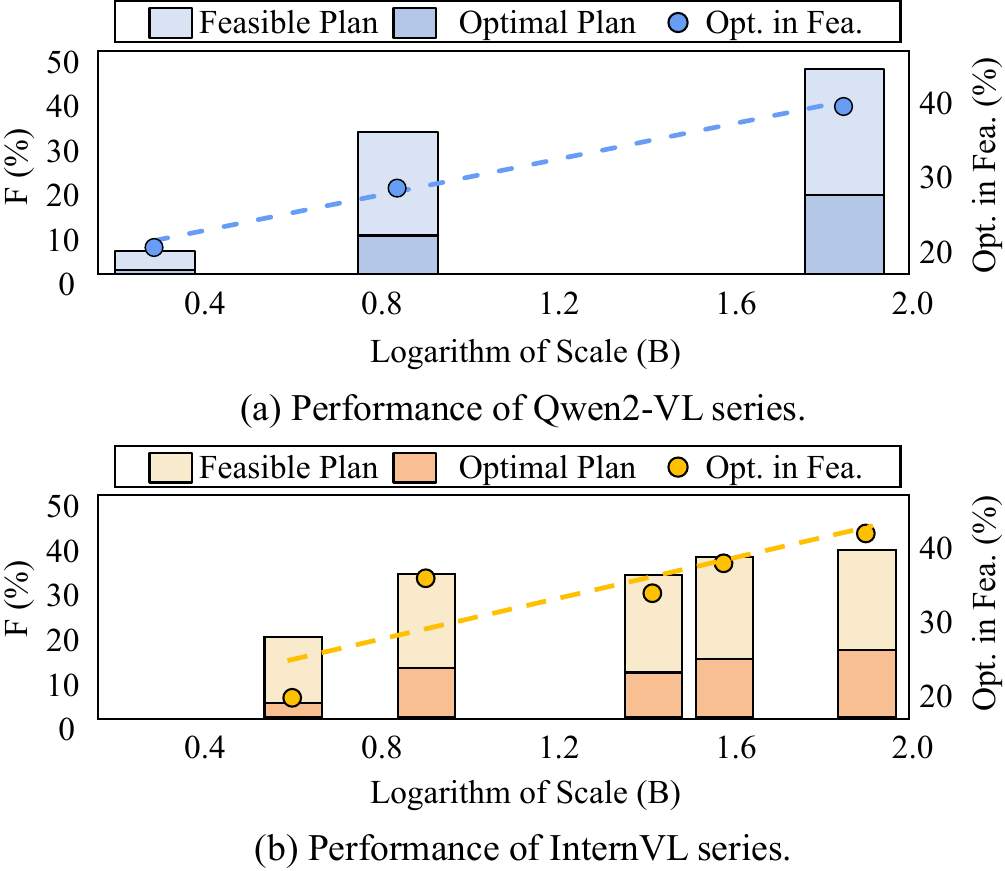}
    \caption{Selected open source model scales and their performance on Flight Planning EASY. \textbf{Opt. in Fea.}: Ratio of \textit{optimal plan} in \textit{feasible plan}.}
    \label{fig:scaling}
\end{figure}

\paragraph{Scaling law can improve the model performance}
A clear trend emerges from Table~\ref{main}: larger models yield better optimal plan performance, with gains notably exceeding those for feasible plans. We analyze the correlation between plan outcomes and model size. Figure~\ref{fig:scaling} shows the ratio of \textit{optimal plans} within \textit{feasible plans} for Flight Planning across models. This suggests that scaling up MLLMs improves their ability to balance constraints and optimization goals, enabling broader solution exploration. Such findings may inform the design of MLLMs that better emulate human planning.

\subsection{Exploration}
\subsubsection{Chain-of-thought prompting strategies are effective for feasible planning but limited in optimal planning}
We examine the effectiveness of prompt strategies such as Chain-of-Thought~\cite{wei2022chain} and Plan-and-Solve~\cite{wang2023plan} in enhancing MLLM planning across tasks in MPCC. As shown in Table~\ref{tab:prompt}, their impact depends on task type and difficulty. These methods are especially useful in simpler scenarios. For instance, in the EASY level of Flight Planning, GPT-4o's accuracy rises from 65.0\% (Direct) to 74.0\% (PS), showing effectiveness in meeting initial constraints. However, as complexity grows—particularly in Calendar and Meeting Planning—the advantage narrows to 2.3\% or becomes negative, indicating that prompting benefits diminish with more complex multimodal constraints.

Conversely, improvements in optimal plan performance are less clear and consistent. While some models benefit from prompt strategies in low-constraint tasks, their ability to find the optimal solution declines as complexity grows. This suggests that although prompt strategies aid basic constraint satisfaction, they fall short of supporting fine-grained optimization in complex multimodal settings.

\begin{figure}[t]
\centering
\includegraphics[width=0.40\textwidth]{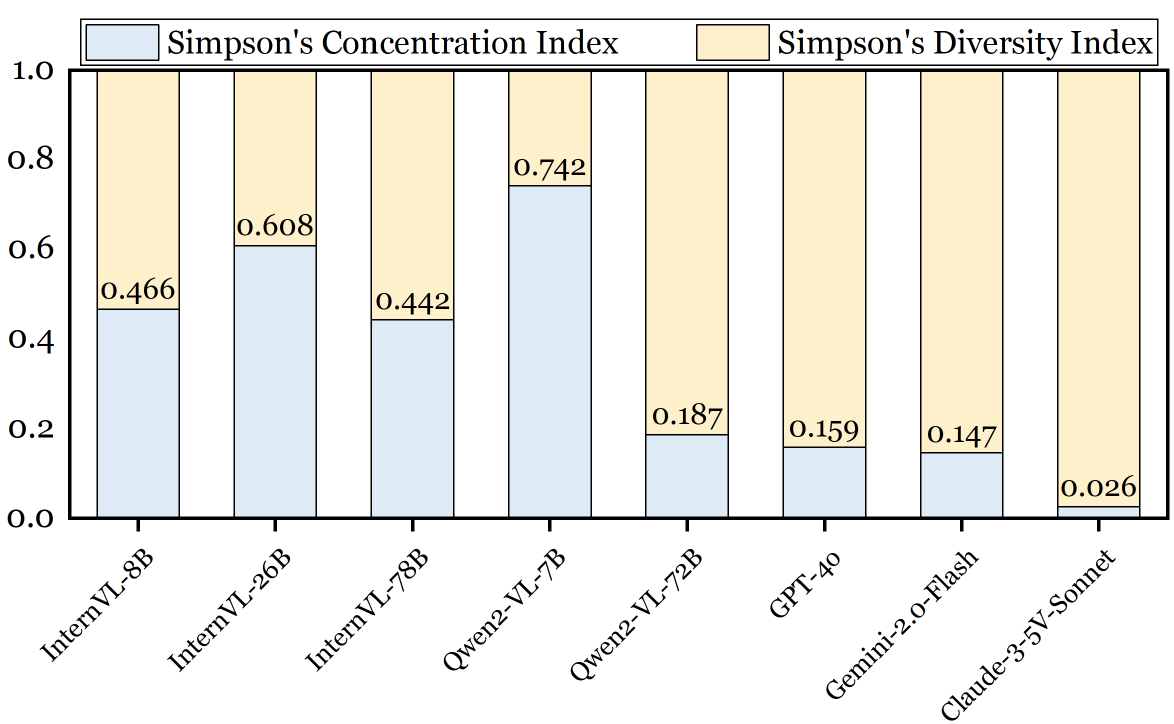}
\caption{Simpson's Concentration and Diversity Index of responses for Meeting Planning HARD tasks.}
\label{fig:bias and model}
\end{figure}

\begin{table*}[t]
	\centering
	\caption{
		Evaluating results under different prompt strategies.
	}
	\begin{adjustbox}{width=0.90\textwidth}
		\begin{tabular}{lcccccccccc}
			\toprule
			\multirow{2}{*}{Model}  & \multicolumn{3}{c}{Flight Planning} & \multicolumn{3}{c}{Calendar Planning} & \multicolumn{3}{c}{Meeting Planning} & \multirow{2}{*}{Average}
			\\\cmidrule{2-10}
			& EASY & MEDIUM & HARD & EASY & MEDIUM & HARD & EASY & MEDIUM & HARD & 
			\\
			\midrule
			Empirical Max & - & - & - & 11.7/10.0 & \textbf{11.3}/\textbf{10.3} & \textbf{7.0}/\textbf{5.3} & 12.0/8.0 & \textbf{8.0}/5.3 & \textbf{7.0}/\textbf{6.3} & - \\
			\midrule
			\rowcolor{gray!8}\multicolumn{11}{c}{\textit{GPT-4o}}\\
			\midrule
			{Direct} & 65.0/33.0 & 16.7/1.7 & 2.7/0.0 & 24.0/21.0 & 9.0/7.7 & 2.0/1.0 & 17.7/16.0 & 7.7/5.3 & 4.0/3.3 & 16.7/9.9 \\
			{CoT} & 66.0/47.0 & 48.7/\textbf{12.0} & \textbf{7.7}/\textbf{1.0} & 29.0/24.7 & 10.0/7.3 & 2.7/0.3 & 16.0/15.0 & 5.3/5.0 & 3.3/3.0 & 21.0/12.8 \\			
            {PS} & 74.0/56.0 & \textbf{50.0}/10.3 & 4.3/0.3 & 26.3/22.3 & 9.3/7.0 & 0.7/0.0 & 16.3/15.0 & 6.7/\textbf{6.0} & 4.3/4.0 & 21.3/13.4 \\
			\midrule
			\rowcolor{gray!8}\multicolumn{11}{c}{\textit{Gemini-2.0-Flash}}\\
			\midrule
            {Direct} & 53.3/27.0 & 11.7/1.0 & 0.7/0.0 & 21.0/12.0 & 5.7/1.7 & 3.3/1.0 & 25.0/23.0 & 6.7/4.3 & 4.0/2.7 & 14.6/8.1 \\
            {CoT} & 71.3/54.3 & 38.3/10.3 & 2.0/0.0 & 27.7/21.3 & 5.0/0.3 & 4.0/3.0 & 26.7/24.3 & 6.3/5.7 & 5.0/4.0 & 20.7/13.7 \\
            {PS} & 69.3/\textbf{61.7} & 37.3/7.7 & 3.0/0.3 & 26.3/20.7 & 8.7/5.3 & 3.7/1.3 & 24.7/23.0 & 6.3/5.3 & 2.0/2.0 & 20.1/14.1 \\
            \midrule
			\rowcolor{gray!8}\multicolumn{11}{c}{\textit{Claude-3.5V-Sonnet}}\\
        \midrule
    	{Direct} & 75.3/46.7 & 34.0/4.7 & 0.0/0.0 & 29.0/23.3 & 9.3/2.0 & 2.0/1.7 & \textbf{37.7}/\textbf{34.3} & 2.7/1.7 & 2.0/1.7 & 21.3/12.9 \\
        {CoT} & 76.7/44.7 & 45.7/7.3 & 0.3/0.0 & \textbf{37.3}/\textbf{30.3} & 6.7/1.3 & 1.7/0.3 & 32.3/28.7 & 5.7/5.0 & 2.7/2.0 & \textbf{23.2}/13.3 \\
        {PS} & \textbf{81.0}/46.0 & 38.7/6.7 & 0.0/0.0 & 29.0/23.7 & 9.3/2.3 & 0.7/0.7 & 36.7/33.7 & 4.3/3.3 & 2.0/1.3 & 22.4/\textbf{14.2} \\
			\bottomrule
		\end{tabular}
	\end{adjustbox}
	
	\label{tab:prompt}
\end{table*}

\subsubsection{Smaller MLLMs exhibit significant planning bias during multimodal planning in MPCC}
\label{sec:bias}
When constrained by Concurrent Coordination, we observe that MLLMs often generate infeasible plans, consistently favoring certain patterns, regardless of input. This reveals a strong response bias. To further examine performance degradation, we quantify this tendency in Meeting Planning HARD tasks using ecological and statistical indices: Simpson's Diversity Index and Concentration Index, which measure response dominance and variety.
As shown in Figure~\ref{fig:bias and model}, models with fewer parameters exhibit greater bias in unsolvable cases. This likely stems from the limited reasoning capacity of smaller models, which struggle with complex problems. Enhancing their ability to plan under intricate multimodal constraints is therefore critical.

\begin{figure}[t]
\centering
\includegraphics[width=0.40\textwidth]{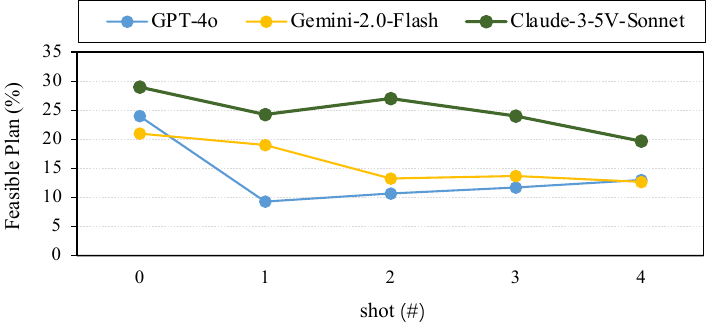}
\caption{In-Context-Learning analysis on \textit{feasible plan} performance with text-only demonstrations.}
\label{fig:icl}
\end{figure}

\subsubsection{The text-only in-context learning approach does not work effectively in MPCC}
In addition to zero-shot evaluations, following Qin~\textit{et al.}~\cite{qin2024factors}, we conducted In-Context Learning experiments on Calendar Planning EASY to explore ways to enhance model planning. Due to the limitations on image input, we use text-only demonstrations to provide contextual information for in-domain tasks.
As shown in Figure~\ref{fig:icl}, ICL negatively impacts most closed-source models. Notably, GPT-4o's performance drops below the Empirical Max. Moreover, for the two currently optimal MLLMs, performance declines significantly as the example shots increase. This suggests the model fails to learn an effective approach for solving multimodal planning using text-only demonstrations.

\begin{figure}[t]
	\centering
	\includegraphics[width=0.40\textwidth]{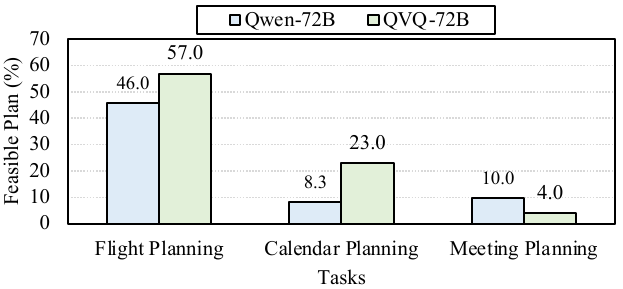}
	\caption{Performance of QvQ-72B-Preview vs. Qwen2-VL-72B at the EASY level for each task.}
	\label{fig:qvq}
	\end{figure}

\subsubsection{Reasoning MLLMs Exploration}
Recent advancements in reasoning MLLMs have shown significant success~\citep{guo2025deepseek,chen2024unlocking,chen2025towards,chen2025ecm}. To assess how reasoning affects multimodal tasks with complex constraints, we evaluated QvQ-72B-Preview~\cite{qvq} on some MPCC tasks and found that: (1) \textbf{Improved reasoning mechanisms enhance performance in simpler multimodal tasks.} As shown in Figure~\ref{fig:qvq}, QvQ outperforms Qwen2-VL-72B in tasks with fewer constraints, despite having the same size of parameters. This suggests that advanced reasoning benefits simpler planning. (2) \textbf{Complex constraints can hinder MLLMs' reasoning.} Performance drops in Meeting Planning with higher constraint complexity, suggesting complexity may induce overthinking or rigid reasoning. Qualitative analysis reveals two main failure modes: over-focusing on specific constraints and premature termination from internal contradiction.

These findings show that reasoning MLLMs excel in low-constraint tasks but struggle with overthinking with complex constraints.

\begin{figure}[t]
\centering
\includegraphics[width=0.40\textwidth]{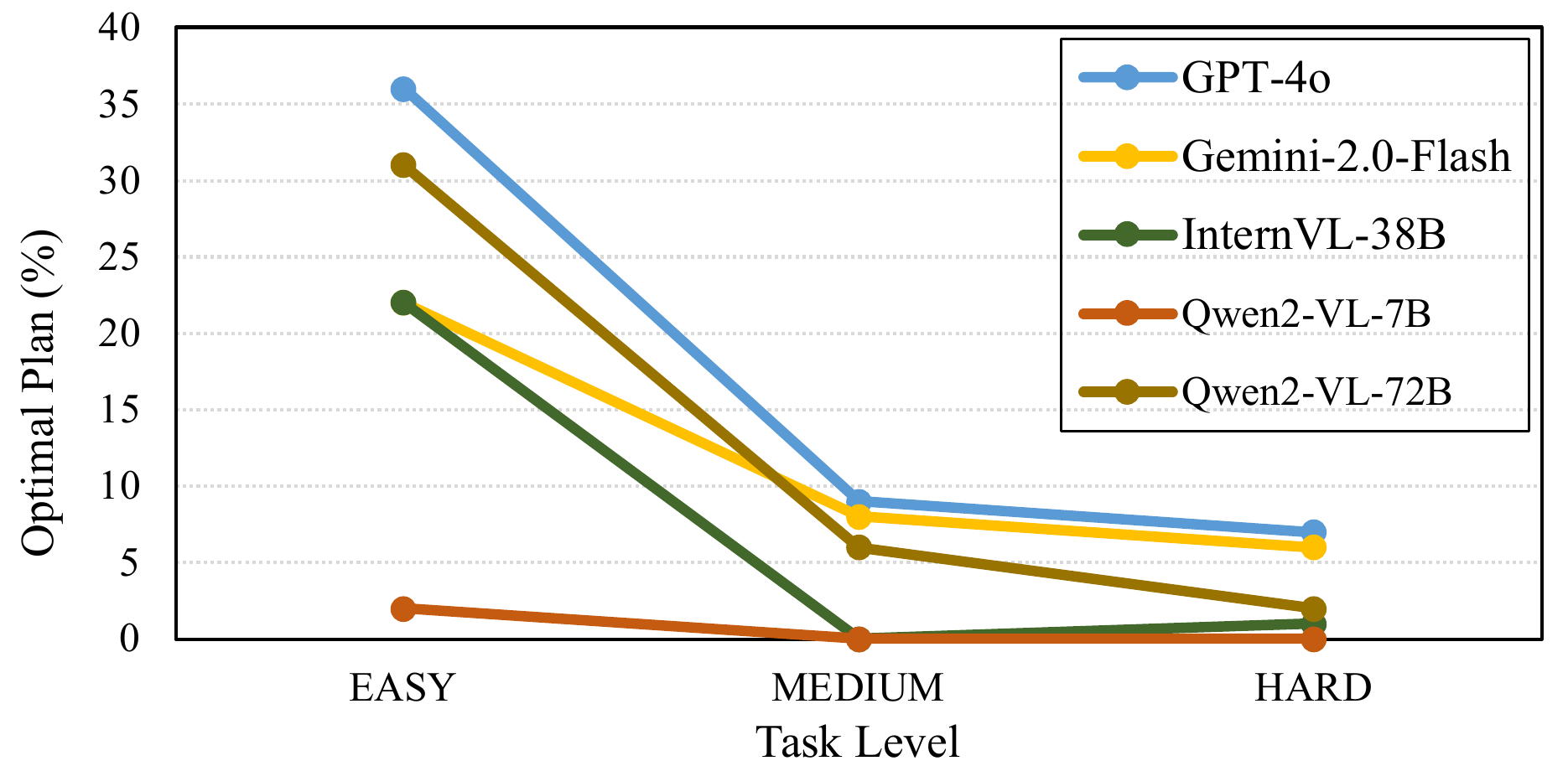}
\caption{Performance of MLLMs on Flight Planning tasks in MPCC, where visual information has been converted into textual input.}
\label{fig:img2text}
\end{figure}

\subsubsection{Both Visual Understanding and Complex Constraints Integration Cause Performance Degradation}

To pinpoint why performance falls from EASY to HARD, we run a control in the Flight Planning tasks by converting visual inputs into structured text to force models to depend solely on language reasoning. As shown in Figure~\ref{fig:img2text}, MLLMs' optimal plan accuracy rises, yet the drop from EASY to HARD remains. We also measured OCR accuracy on Flight Planning images and found that Gemini-2.0-Flash reaches 98.8\% on HARD level, yet complex planning remains challenging, indicating that both visual comprehension and constraint integration drive the decline. Addressing one in isolation is unlikely to close the gap.

\subsection{Case Studies}
\label{sec:case_study}

This section presents a case study analyzing failures of the Claude-3.5V-Sonnet model with Chain-of-Thought~\cite{wei2022chain}, to identify common error causes and specific MLLM weaknesses. Errors are categorized as follows:
(1) \textbf{\textcolor{E}{Information Extraction Error}}: Failures in extracting key visual details, such as misreading flight info or dates.
(2) \textbf{\textcolor{L}{Linguistic Logic Confusion}}: Responses containing contradictions or logical inconsistencies. (3) \textbf{\textcolor{S}{Incomplete Program Search}}: Prematurely concluding no solution exists when further exploration might find one. (4) \textbf{\textcolor{F}{Output Format Error}}: Correct responses that fail to follow the expected output format. (5) \textbf{\textcolor{C}{Violation of Constraints}}: Critical errors where responses, though logically sound, fail to meet the constraints.

\begin{figure}[t]
\centering
\includegraphics[width=0.46\textwidth]{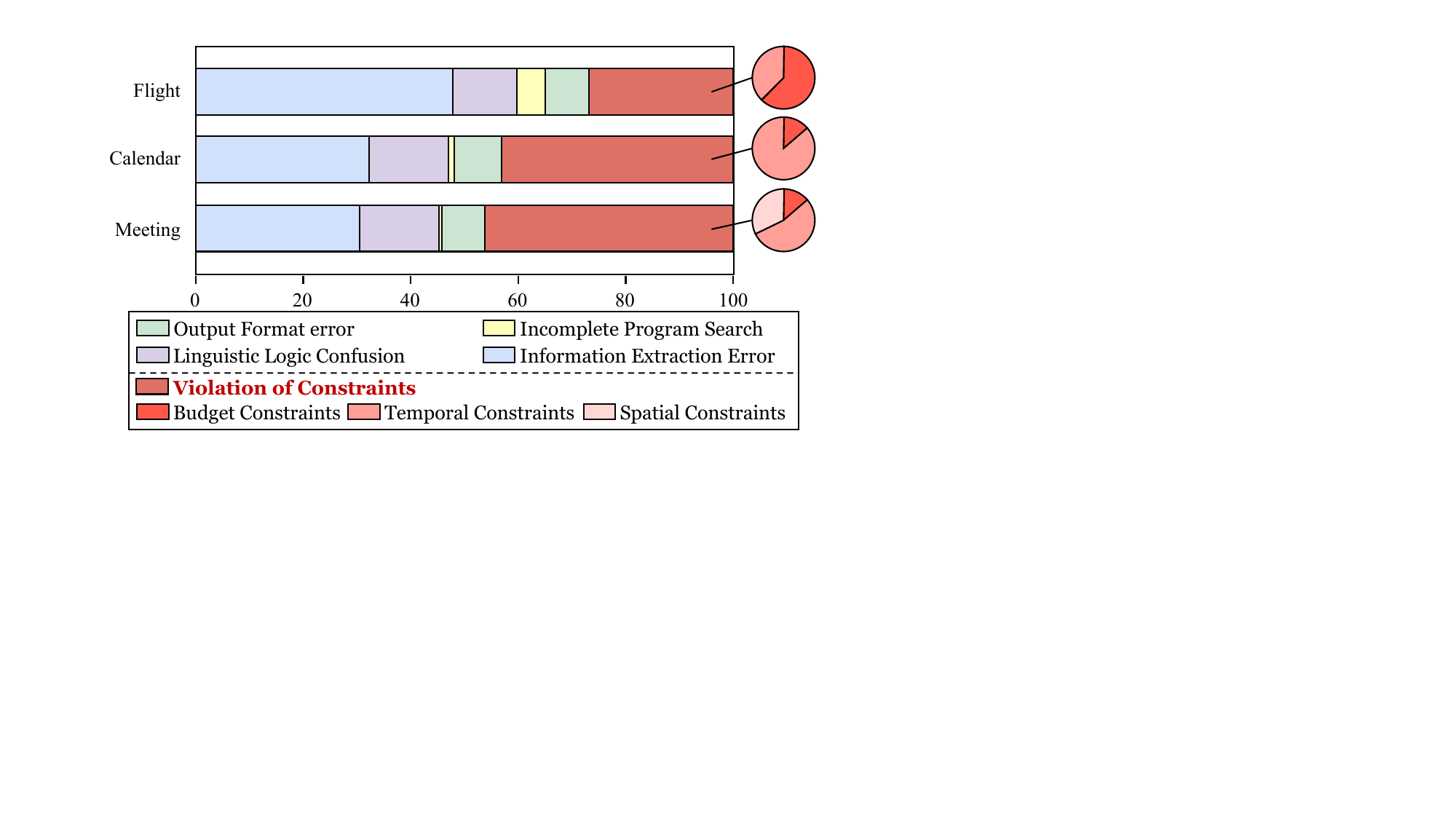}
\caption{Manual analysis of incorrect responses. Over 40\% of incorrect responses were due to failure to meet constraints.}
\label{fig:case}
\end{figure}

As shown in Figure~\ref{fig:case}, over 40\% of errors result from Violation of Constraints, with this percentage rising in tasks with more complex constraints. This highlights the challenge of meeting diverse and complex constraints as the main factor limiting MLLMs' performance in multimodal planning tasks.

\section{Related Work}
The advancement of multimodal large language models (MLLMs) has led to numerous benchmarks evaluating performance on realistic tasks with multimodal constraints. This section reviews existing benchmarks focused on multimodal reasoning.
MMMU~\cite{yue2024mmmu} tests cross-disciplinary problem-solving with 11.5K expert-level questions, revealing significant gaps between MLLMs and human performance. SEED-Bench~\cite{li2023seed} uses a hierarchical evaluation framework with annotated multiple-choice questions across 27 to 34 dimensions. Benchmarks like MLLM-CompBench~\cite{kil2024mllm} and M$^3$CoT~\cite{chen2024m}, refine multimodal reasoning tasks, offer insights into the limitations of MLLMs and drive progress in model development. XLRS-Bench~\cite{wang2025xlrsbenchmultimodalllmsunderstand} introduces complex spatial relationships in high-resolution remote sensing images, challenging MLLMs' reasoning capabilities.
Some benchmarks focus on multimodal planning with implicit constraints. Open3DVQA~\cite{zhan2025open3dvqabenchmarkcomprehensivespatial} tests spatial planning by implicitly restricting spatial relationships. EgoPlan-Bench~\cite{chen2023egoplan} constrains models to first-person perspectives, while VisualWebArena~\cite{koh2024visualwebarena} and OSWorld~\cite{xie2024osworld} involve interacting with real-world interfaces and following specific instructions through embodied planning.

\section{Conclusion}
This study introduces the Multimodal Planning with Complex Constraints (MPCC) benchmark, designed to evaluate multimodal large language models (MLLMs) on real-world planning tasks with complex multimodal constraints. Experiments with 13 state-of-the-art MLLMs revealed significant limitations: even advanced models like Claude-3.5V-Sonnet struggle with intricate constraints, and open-source models show even greater deficiencies. We also explored the application of multiple strategies, and the results highlight the shortcomings of current prompting strategies in handling multiple constraints and emphasize the need for further development in constraint-aware reasoning within MLLMs. MPCC provides a robust framework for systematically evaluating MLLMs under diverse real-world constraints to guide future research toward more capable and reliable multimodal planning systems.

\begin{acks}
This work was supported by the National Natural Science Foundation of China (NSFC) via grant 62306342, 62236004, 62206078 and 62476073. This work was supported by the Scientific Research Fund of Hunan Provincial Education Department (24B0001). This work was sponsored by the Excellent Young Scientists Fund in Hunan Province (2024JJ4070), the Science and Technology Innovation Program of Hunan Province under Grant 2024RC3024 and CCF-Zhipu Large Model Innovation Fund (NO.CCF-Zhipu202406).
\end{acks}

\bibliographystyle{plain}
\bibliography{main}

\end{document}